\documentclass[10pt,journal,compsoc]{IEEEtran}

\usepackage[utf8]{inputenc} 
\usepackage[T1]{fontenc}    
\usepackage{amsmath}
\usepackage{bbm}
\usepackage{multirow}
\usepackage{makecell}
\usepackage{caption}
\usepackage{tabto}
\usepackage{graphicx}
\usepackage{float}
\usepackage{subfigure}
\usepackage{hyperref}
\usepackage{enumitem}
\usepackage{cite}
\usepackage{enumitem}

\begin{document}

\title{Cross-view Self-Supervised Learning on \\ Heterogeneous Graph Neural Network \\ via Bootstrapping}

\author{Minjae~Park
    \IEEEcompsocitemizethanks{
        \IEEEcompsocthanksitem Minjae Park is with the School of Computing, Korea Advanced Institute of Science and Technology, Daejeon, Republic of Korea; e-mail: mingyeonri@gmail.com
    }
}

\IEEEtitleabstractindextext{%
\begin{abstract}
Heterogeneous graph neural networks can represent information of heterogeneous graphs with excellent ability. Recently, self-supervised learning manner is researched which learns the unique expression of a graph through a contrastive learning method. In the absence of labels, this learning methods show great potential. However, contrastive learning relies heavily on positive and negative pairs, and generating high-quality pairs from heterogeneous graphs is difficult. In this paper, in line with recent innovations in self-supervised learning called BYOL or bootstrapping, we introduce a  that can generate good representations without generating large number of pairs. In addition, paying attention to the fact that heterogeneous graphs can be viewed from two perspectives, network schema and meta-path views, high-level expressions in the graphs are captured and expressed. The proposed model showed state-of-the-art performance than other methods in various real world datasets.
\end{abstract}

}

\maketitle

\IEEEdisplaynontitleabstractindextext

\IEEEraisesectionheading{\section{Introduction}\label{sec:introduction}}
\IEEEPARstart{I}{n} real world, heterogeneous graph\cite{DBLP:journals/sigkdd/SunH12} models various kinds of nodes and their relationships, such as bibliographic networks\cite{hu2020strategies} or movie  networks. The heterogeneous graph neural network effectively represents various relationships between these heterogeneous nodes, resulting in performance improvements in various sections, such as recommendation systems \cite{herec}.

As performed in the existing general homogeneous graph neural networks\cite{gcn, gat, GraphSAGE}, heterogeneous graph neural networks also belong to semi-supervised learning\cite{han,  magnn}. Although labeling of the entire node is not required, sometimes such labeling is difficult. For example, classification of papers is difficult for non-experts. To solve this problem, self-supervised learning method was introduced\cite{cpc,moco,simclr,dgi,mvgrl}. The most used learning method in self-supervised learning is contrastive learning. Contrastive learning requires positive and negative pairs, and during training place embeddings of positive pairs close to each other and embeddings of negative pairs farther away. However, implementing contrastive learning on heterogeneous graphs is not an easy task.

First, high-quality contrastive learning requires a large amount of positive and negative pairs. However, in heterogeneous graphs, it is difficult to create many of these high-quality positive or negative pairs. In a typical graph, we expect neighboring nodes to have the same properties, so we can produce positive or negative pairs depending on whether they are neighbors or not. However, in heterogeneous graphs, this method may produce incorrect pairs. This is because in heterogeneous graphs, adjacent nodes with different meanings exist. For example, in a movie network, it is difficult to expect movie nodes connected via actor nodes to have the same characteristics. This is usually because actors appear in films of different characteristics. A recent study proposed BGRL\cite{bgrl}, which improved the problem of contrastive learning. This method has been shown to enable self-supervised learning without generating large numbers of pairs, especially negative pairs.

However, this approach introduces a second problem. Since there are many non-attribute graphs in heterogeneous graphs, when augmentation of the existing attribute masking method is used, even if BGRL is used, the maximum performance is not achieved. In order to solve this problem, a new augmentation method has been proposed\cite{heco} recently, focusing on the fact that heterogeneous graphs can be viewed from two perspectives. However, these studies require two models and cannot take advantage of the BGRL. In this study, by combining the two methods, we propose a model that can express heterogeneous graphs from two perspectives while sharing trainable parameters.

\begin{itemize}
    \item To the best of our knowledge, this is the first attempt to study more complex learning in BYOL in heterogeneous graphs.
    \item Our model can generate two-view node representations with a single model, which shows improved performance even in non-attribute graphs where attribute masking is useless.
    \item We perform various experiments on four real-world datasets and show that the proposed model achieved the best performance in various respects.
\end{itemize}

\section{Related Works}
\subsection{Heterogeneous Graph Network}
A heterogeneous graph neural network is a neural network for effectively representing a heterogeneous graph consisting of various types of nodes and links. HAN\cite{han} uses attention at the node level and semantic level. MAGNN\cite{magnn} references to intermediate nodes between nodes that are not considered in HAN through various techniques. For GTN\cite{gtn}, it automatically catches and uses useful connections rather than predefined ones. HGT\cite{hgt} also automatically captures meaningful connections and can be easily scaled up for large datasets. HetGNN\cite{hetegnn} samples neighbors and uses them after encoding them into LSTM. However, these neural networks do not attempt self-supervised learning.

\subsection{Contrastive learning}
Contrastive learning refers to learning by placing embeddings of positive pairs close and embeddings of negative pairs far away, based on positive and negative pairs. In DGI\cite{dgi}, we learn by maximizing the amount of mutual information by using the local and global embeddings generated by the normal graph as positive pairs and the local embeddings generated by the corrupt graph and global embeddings generated by the normal graph as negative pairs. GMI\cite{gmi} learns by capturing a central node within the topology and placing this node and local embeddings in positive pairs. GCC places the local embeddings of the two graphs in positive pairs. DMGI\cite{dmgi} conduct contrastive learning on the normal graph and the corrupt graph through the meta-path. In HeCo\cite{heco}, it use different views to create positive and negative pairs through metapaths and then train them in contrast. Either they did not fully utilize the characteristics of heterogeneous graphs, or they could not produce positive or negative pairs properly.

\subsection{Bootstrap your own latent}
To solve the problem of contrastive learning, which requires generating a large amount of positive and negative pairs, BGRL\cite{bgrl} has been proposed which depending on BYOL method. Self-supervised learning is performed using augmentation methods based on node masking or edge masking, and two encoders and one predictor. This method enables self-supervised learning with only positive pairs. In addition, AFGRL\cite{afgrl}, an additional BYOL-based self-supervised learning method that does not use augmentation, has been proposed. However, in the case of BGRL, performance is poor in non-attribute graph due to the nature of reliance on masking, and in the case of AFGRL, it is difficult to reflect various types of nodes in training.

\section{Preliminary}
\subsection{Heterogeneous Graph}
Heterogeneous graph is defined as a $\mathcal{G}=(\mathcal{V},\mathcal{E},\mathcal{A},\mathcal{R},\mathcal{\phi},\mathcal{\varphi})$. $\mathcal{V}$ and $\mathcal{E}$ denote sets of nodes and edges. $\mathcal{A}$ and $\mathcal{R}$ denotes sets of object and link types and $|\mathcal{A}+\mathcal{R}|>2$. Finally, there are mapping functions $\mathcal{\phi}:\mathcal{V}\rightarrow\mathcal{A}$ and $\mathcal{\varphi}:\mathcal{E}\rightarrow\mathcal{R}$, which denotes for node mapping function and edge mapping function respectively.

\subsection{Network Schema}
The network schema denoted by $T_G=(\mathcal{A},\mathcal{R})$ represents the Heterogeneous graph $\mathcal{G}$ as a meta-template. The object $\mathcal{A}$ is connected as a directed graph through the link type $\mathcal{R}$. 

\subsection{Meta-path}
A meta-path $\mathcal{P}$ is defined as a path in the form of $A_1\stackrel{R_1}{\longrightarrow}A_2\stackrel{R_2}{\longrightarrow}\dots\stackrel{R_l}{\longrightarrow}A_{l+1}$ (abbreviated as $A_1A_2\dots A_{l+1}$), which represents a composite relation $R=R_1\circ R_2\circ \dots\circ R_l$ between node types $A_1$ and $A_{l+1}$, where $\circ$ denotes the relations composition operator.

\section{Proposed Model: CSGRL}
In this section, we introduce our model that encodes the nodes and then trains them through the BYOL method. As a single model, our model can represent nodes through two views: a network schema view and a meta-path view. After that, self-supervised learning is performed on two views of the same node through the BYOL method.

\subsection{Encoder}

\begin{figure}[H]
\centering
\captionsetup{justification=centering}
\includegraphics[width=\linewidth]{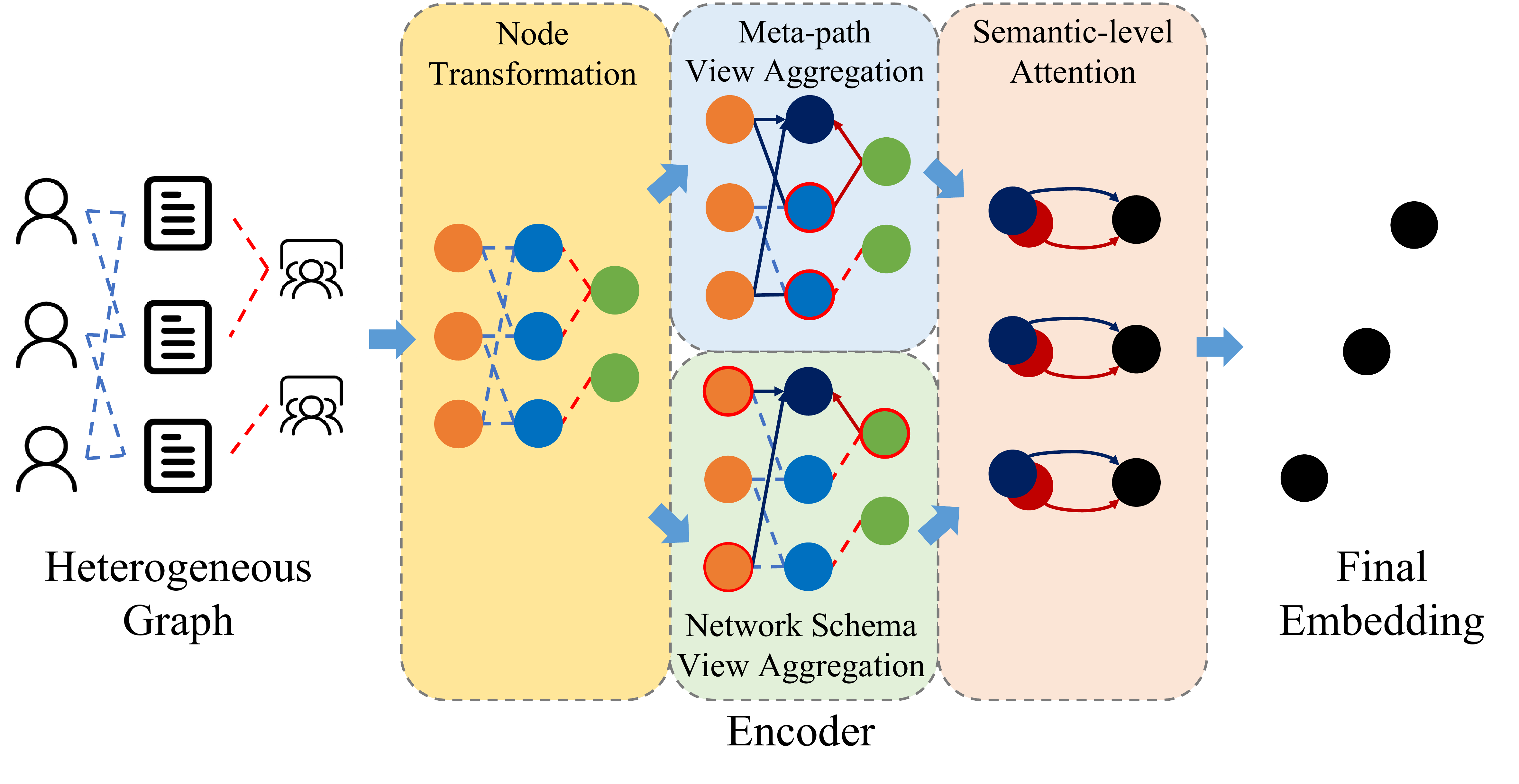}
\caption{The structure of Encoder}
\label{fig:encoderprocess}
\end{figure}

\subsubsection{Node Feature Transformation}
Since there are different types of nodes, we need to transform the node features in different dimension into same dimension. Converts the feature $x_i$ of the node $i$ with type $\phi_i$ into $h_i$ through each type conversion parameter, matrix $W_{\phi_i}$ and bias $b_{\phi_i}$.
\begin{equation}
\label{eq:node_transfrom}
    h_i = \sigma\left(W_{\phi_i}\cdot x_i + b_{\phi_i}\right)
\end{equation}

\subsubsection{Network Schema View Aggregation}
In the network schema view, node $i$ connects to nodes of $M$ different types $\{ \Phi_1, \Phi_2,...,\Phi_M \}$. Among them, the neighbor node connected through the $\Phi_m$ type is defined as $N^{\Phi_m}_i$. For example, paper ${P_1}$ is connected with neighbor nodes $N^{\Phi_A}_{P_1}=\{A_1, A_2, ..., A_n\}$ of author type $\Phi_A$ and neighbor nodes $N^{\Phi_S}_{P_1}=\{S_1, S_2, ..., S_n\}$ of subject type $\Phi_S$. Aggregate by applying the RGCN\cite{rgcn} aggregation method. The embedding $h^{\Phi_m}_i$ created by aggregating the $\Phi_m$ type neighbors $N^{\Phi_m}_i$ of the node $i$ is defined as:
\begin{equation}
\label{eq:networkschemeaviewaggr}
    h^{\Phi_m}_i = \sigma\left(\frac{1}{|N^{\Phi_m}_i|}\sum\limits_{j \in N^{\Phi_m}_i}{h_j}\right)
\end{equation}

\subsubsection{Meta-path View Aggregation}
The meta-path view aggregates neighbors connected via meta-path. For $i$ nodes, there is a $N$ meta-paths $\{ \mathcal{P}_1, \mathcal{P}_2,...,\mathcal{P}_N \}$. A neighbor node of $i$ connected through each meta path $\mathcal{P}_n$ is defined as $N^{\mathcal{P}_n}_i$. For example, if paper ${P_1}$ is connected to $P_2$ via author $A_1$, $P_2$ is one of the neighbor nodes $N^{\mathcal{P}_1}_{P_1}$ of ${P_1}$ connected via meta-path $\mathcal{P}_1$(PAP). The GCN aggregation method is used to aggregate these neighbor nodes. The embedding $h^{\mathcal{P}_n}_i$ created by aggregating $N^{\mathcal{P}_n}_i$ and node $i$ is defined as:
\begin{equation}
\label{eq:metapathviewaggr}
    h^{\mathcal{P}_n}_i = \sigma\left(\frac{1}{d_i+1}h_i + \sum\limits_{j \in N^{\mathcal{P}_n}_i}\frac{1}{\sqrt{(d_i+1)(d_j+1)}}h_j\right)
\end{equation}

\subsubsection{Semantic-level Attention}
We aggregate the embedding $h_i$ generated for each meta-path $\mathcal{P}_n$ or node type $\Phi_m$ of node $i$ into one final embedding $z_i$ through attention. final embedding $z_i$ is defined as:
\begin{equation}
\label{eq:finalembedding}
    z_i = \sum\limits^T_{t=1}\beta_{\mathcal{T}_t} \cdot h^{\mathcal{T}_t}_i
\end{equation}
If meta-path view aggregate is performed, $T$ becomes $N$, view type $\mathcal{T}$ becomes $\mathcal{P}$, and when network scheme view aggregate is performed, $T$ becomes $M$ and view type $\mathcal{T}$ becomes $\Phi$. The final embedding generated through the meta-path view aggregate becomes $z^{mp}_i$, and the final embedding generated through the network scheme view aggregate becomes $z^{sc}_i$.
$\beta_\mathcal{T}$ denotes importance of view type $\mathcal{T}$. $\beta_\mathcal{T}$ is defined as
\begin{equation}
\label{eq:finalembedding_att}
\begin{split}
    w_{\mathcal{T}_t} &= \frac{1}{|V|} \sum\limits_{i \in V} a^{\top} \cdot \text{tanh}\left(Wh^{\mathcal{T}_t}_i+b\right), \\
    \beta_{\mathcal{T}_t} &= \frac{\text{exp}(w_{\mathcal{T}_t})}{\sum^T_{i=1}\text{exp}(w_{\mathcal{T}_i})}
\end{split}
\end{equation}

\subsection{Bootstrap your own latent}

\begin{figure}[H]
\centering
\captionsetup{justification=centering}
\includegraphics[width=\linewidth]{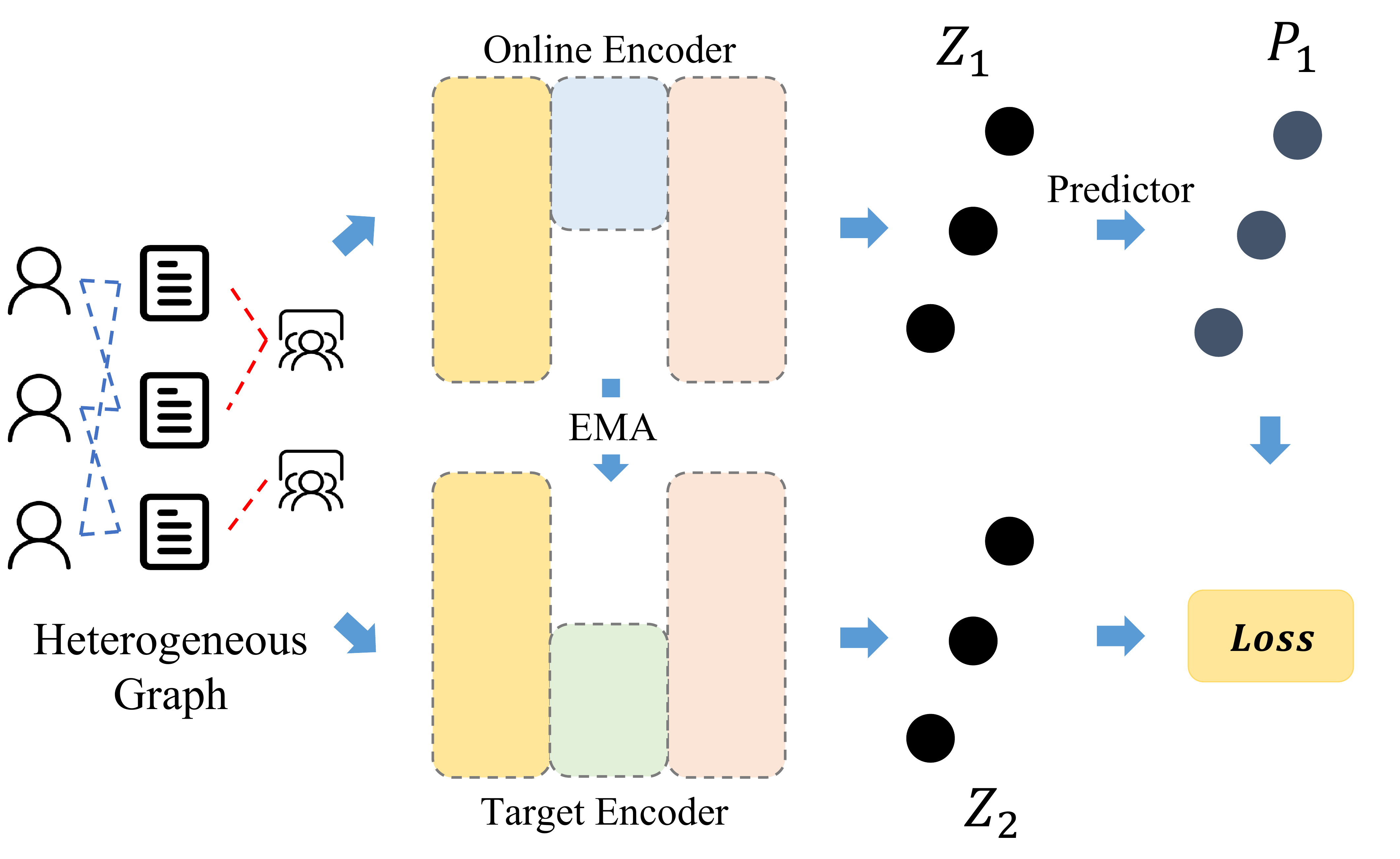}
\caption{The structure of BYOL process when online encoder use meta-path view aggregation}
\label{fig:byolprocess}
\end{figure}

\subsubsection{CSGRL Component}
Prepare two encoders for CSGRL training. These are the online encoder $\mathcal{E}_\theta$ and a target encoder $\mathcal{E}_\pi$, respectively. Also, prepare the heterogeneous graph $\mathcal{G}$. After that, we use the online encoder and the target encoder to find $Z_1=\mathcal{E}_\theta(\mathcal{G})$ and $Z_2=\mathcal{E}_\pi(\mathcal{G})$. We also use the predictors $p_\theta$ to find $P_1=p_\theta(Z_1)$. In this process, $Z_1$ and $Z_2$ must be created using different view aggregate methods. That is, if $Z_1$ is $Z^{sc}$, then $Z_2$ must be $Z^{mp}$, and when $Z_1$ is $Z^{mp}$, $Z_2$ must be $Z^{sc}$. 

\subsubsection{Finding positive neighbors}
To generate a small but high-quality positive pair, we create a set of positive neighbors $\mathcal{N}_i$ for each node $i$. $\mathcal{N}_i$ is a subset of the set of all neighboring nodes $N^{\mathcal{P}}_i= \bigcup^{N}_{n=1}N^{\mathcal{P}_n}_i$ connected via meta-path. Subset $\mathcal{N}_i$ is created taking into count the number of meta-paths connected in $N^{\mathcal{P}}_i$. For example, add to $\mathcal{N}_{v_1}$ the neighbor node $v_3$ of $v_1$, which belongs to both $N^{\mathcal{P}_1}_{v_1}$ and $N^{\mathcal{P}_2}_{v_1}$, rather than the neighbor node $v_2$ of $v_1$ that belongs only to $N^{\mathcal{P}_1}_{v_1}$. 

\subsubsection{Updating the online encoder}
The online parameter $\theta$ (not $\pi$) is updated so that the predicted $P_1$ is similar to $Z_2$ via the gradient computed via cosine similarity.
\begin{equation}
\label{eq:bgrl_online_update}
    \ell(\theta, \pi) = -\frac{2}{|V|}\sum\limits_{i=1}^{|V|}\sum\limits_{j \in \mathcal{N}_i}\frac{P_{(1, i)} Z_{(2, j)}^{\top}}{\|P_{(1, i)}\| \|Z_{(2, j)}\|}
\end{equation}
\begin{equation}
\label{eq:bgrl_online_update_optim}
\theta \leftarrow \text{optimize}(\theta,~\eta,~\partial_\theta \ell(\theta, \pi)),
\end{equation}
$\eta$ is the learning rate, and only $\theta$ is updated through the calculated $\ell$. And we also symmetrize this loss by using the online representation of the second view to also predict the target representation of the first view. 

\subsubsection{Updating the target encoder}
The target parameter $\pi$ is updated as an exponential moving average of the online parameter $\theta$ using the damping factor $\tau$.
\begin{equation}
\pi \leftarrow \tau \pi + (1 - \tau) \theta,
\end{equation}
\section{Experiment}
\subsection{Experimental Setup}
\begin{table}[H]
  \caption{The information of the datasets}
  \label{tab:statistics}
  \setlength{\tabcolsep}{3mm}{
  \begin{tabular}{|c|c|c|c|}
    \hline
        Dataset & Node & Relation & Meta-path\\
    \hline
    ACM & \makecell*[c]{paper (P):4019\\author (A):7167\\subject (S):60} & \makecell*[c]{P-A:13407\\P-S:4019} & \makecell*[c]{PAP\\PSP} \\
    \hline
    DBLP & \makecell*[c]{author (A):4057\\paper (P):14328\\conference (C):20\\term (T):7723} & \makecell*[c]{P-A:19645\\P-C:14328\\P-T:85810} & \makecell*[c]{APA\\APCPA\\APTPA} \\
    \hline
    Freebase & \makecell*[c]{movie (M):3492\\actor (A):33401\\direct (D):2502\\writer (W):4459} & \makecell*[c]{M-A:65341\\M-D:3762\\M-W:6414} & \makecell*[c]{MAM\\MDM\\MWM} \\
    \hline
    AMiner & \makecell*[c]{paper (P):6564\\author (A):13329\\reference (R):35890} & \makecell*[c]{P-A:18007\\P-R:58831} & \makecell*[c]{PAP\\PRP} \\
    \hline
\end{tabular}}
\end{table}

\begin{table*}
  \caption{Quantitative results (\%$\pm\sigma$) on node classification.}
  \label{tab:result1}
  \resizebox{\textwidth}{!}{
  \begin{tabular}{c|c|c|ccccccccc|c}
    \hline
    Datasets & Metric & Split & GraphSAGE & GAE & Mp2vec & HERec & HetGNN & HAN & DGI & DMGI & HeCo & CSGRL \\
    \hline
    \multirow{9}{*}{ACM}&
    \multirow{3}{*}{Ma-F1}
    &20&47.13$\pm$4.7&62.72$\pm$3.1&51.91$\pm$0.9&55.13$\pm$1.5&72.11$\pm$0.9&85.66$\pm$2.1&79.27$\pm$3.8&87.86$\pm$0.2&88.56$\pm$0.8&\textbf{90.49$\pm$0.1}\\
    &&40&55.96$\pm$6.8&61.61$\pm$3.2&62.41$\pm$0.6&61.21$\pm$0.8&72.02$\pm$0.4&87.47$\pm$1.1&80.23$\pm$3.3&86.23$\pm$0.8&87.61$\pm$0.5&\textbf{90.41$\pm$0.4}\\
    &&60&56.59$\pm$5.7&61.67$\pm$2.9&61.13$\pm$0.4&64.35$\pm$0.8&74.33$\pm$0.6&88.41$\pm$1.1&80.03$\pm$3.3&87.97$\pm$0.4&89.04$\pm$0.5&\textbf{90.92$\pm$0.5}\\
    \cline{2-13}
    &\multirow{3}{*}{Mi-F1}
    &20&49.72$\pm$5.5&68.02$\pm$1.9&53.13$\pm$0.9&57.47$\pm$1.5&71.89$\pm$1.1&85.11$\pm$2.2&79.63$\pm$3.5&87.60$\pm$0.8&88.13$\pm$0.8&\textbf{90.28$\pm$0.1}\\
    &&40&60.98$\pm$3.5&66.38$\pm$1.9&64.43$\pm$0.6&62.62$\pm$0.9&74.46$\pm$0.8&87.21$\pm$1.2&80.41$\pm$3.0&86.02$\pm$0.9&87.45$\pm$0.5&\textbf{90.34$\pm$0.4}\\
    &&60&60.72$\pm$4.3&65.71$\pm$2.2&62.72$\pm$0.3&65.15$\pm$0.9&76.08$\pm$0.7&88.10$\pm$1.2&80.15$\pm$3.2&87.82$\pm$0.5&88.71$\pm$0.5&\textbf{90.78$\pm$0.5}\\
    \cline{2-13}
    &\multirow{3}{*}{AUC}
    &20&65.88$\pm$3.7&79.50$\pm$2.4&71.66$\pm$0.7&75.44$\pm$1.3&84.36$\pm$1.0&93.47$\pm$1.5&91.47$\pm$2.3&96.72$\pm$0.3&96.49$\pm$0.3&\textbf{97.36$\pm$0.1}\\
    &&40&71.06$\pm$5.2&79.14$\pm$2.5&80.48$\pm$0.4&79.84$\pm$0.5&85.01$\pm$0.6&94.84$\pm$0.9&91.52$\pm$2.3&96.35$\pm$0.3&96.40$\pm$0.4&\textbf{97.43$\pm$0.1}\\
    &&60&70.45$\pm$6.2&77.90$\pm$2.8&79.33$\pm$0.4&81.64$\pm$0.7&87.64$\pm$0.7&94.68$\pm$1.4&91.41$\pm$1.9&96.79$\pm$0.2&96.55$\pm$0.3&\textbf{97.08$\pm$0.5}\\
    \hline
    \multirow{9}{*}{DBLP}&
    \multirow{3}{*}{Ma-F1}
    &20&71.97$\pm$8.4&90.90$\pm$0.1&88.98$\pm$0.2&89.57$\pm$0.4&89.51$\pm$1.1&89.31$\pm$0.9&87.93$\pm$2.4&89.94$\pm$0.4&91.28$\pm$0.2&\textbf{92.56$\pm$0.1}\\
    &&40&73.69$\pm$8.4&89.60$\pm$0.3&88.68$\pm$0.2&89.73$\pm$0.4&88.61$\pm$0.8&88.87$\pm$1.0&88.62$\pm$0.6&89.25$\pm$0.4&90.34$\pm$0.3&\textbf{91.69$\pm$0.1}\\
    &&60&73.86$\pm$8.1&90.08$\pm$0.2&90.25$\pm$0.1&90.18$\pm$0.3&89.56$\pm$0.5&89.20$\pm$0.8&89.19$\pm$0.9&89.46$\pm$0.6&90.64$\pm$0.3&\textbf{92.20$\pm$0.1}\\
    \cline{2-13}
    &\multirow{3}{*}{Mi-F1}
    &20&71.44$\pm$8.7&91.55$\pm$0.1&89.67$\pm$0.1&90.24$\pm$0.4&90.11$\pm$1.0&90.16$\pm$0.9&88.72$\pm$2.6&90.78$\pm$0.3&91.97$\pm$0.2&\textbf{93.10$\pm$0.1}\\
    &&40&73.61$\pm$8.6&90.00$\pm$0.3&89.14$\pm$0.2&90.15$\pm$0.4&89.03$\pm$0.7&89.47$\pm$0.9&89.22$\pm$0.5&89.92$\pm$0.4&90.76$\pm$0.3&\textbf{92.06$\pm$0.1}\\
    &&60&74.05$\pm$8.3&90.95$\pm$0.2&91.17$\pm$0.1&91.01$\pm$0.3&90.43$\pm$0.6&90.34$\pm$0.8&90.35$\pm$0.8&90.66$\pm$0.5&91.59$\pm$0.2&\textbf{93.09$\pm$0.1}\\
    \cline{2-13}
    &\multirow{3}{*}{AUC}
    &20&90.59$\pm$4.3&98.15$\pm$0.1&97.69$\pm$0.0&98.21$\pm$0.2&97.96$\pm$0.4&98.07$\pm$0.6&96.99$\pm$1.4&97.75$\pm$0.3&98.32$\pm$0.1&\textbf{98.75$\pm$0.1}\\
    &&40&91.42$\pm$4.0&97.85$\pm$0.1&97.08$\pm$0.0&97.93$\pm$0.1&97.70$\pm$0.3&97.48$\pm$0.6&97.12$\pm$0.4&97.23$\pm$0.2&98.06$\pm$0.1&\textbf{98.68$\pm$0.1}\\
    &&60&91.73$\pm$3.8&98.37$\pm$0.1&98.00$\pm$0.0&98.49$\pm$0.1&97.97$\pm$0.2&97.96$\pm$0.5&97.76$\pm$0.5&97.72$\pm$0.4&98.59$\pm$0.1&\textbf{99.07$\pm$0.1}\\
    \hline
    \multirow{9}{*}{Freebase}&
    \multirow{3}{*}{Ma-F1}
    &20&45.14$\pm$4.5&53.81$\pm$0.6&53.96$\pm$0.7&55.78$\pm$0.5&52.72$\pm$1.0&53.16$\pm$2.8&54.90$\pm$0.7&55.79$\pm$0.9&59.23$\pm$0.7&\textbf{59.56$\pm$0.4}\\
    &&40&44.88$\pm$4.1&52.44$\pm$2.3&57.80$\pm$1.1&59.28$\pm$0.6&48.57$\pm$0.5&59.63$\pm$2.3&53.40$\pm$1.4&49.88$\pm$1.9&61.19$\pm$0.6&\textbf{63.34$\pm$0.2}\\
    &&60&45.16$\pm$3.1&50.65$\pm$0.4&55.94$\pm$0.7&56.50$\pm$0.4&52.37$\pm$0.8&56.77$\pm$1.7&53.81$\pm$1.1&52.10$\pm$0.7&60.13$\pm$1.3&\textbf{62.37$\pm$0.5}\\
    \cline{2-13}
    &\multirow{3}{*}{Mi-F1}
    &20&54.83$\pm$3.0&55.20$\pm$0.7&56.23$\pm$0.8&57.92$\pm$0.5&56.85$\pm$0.9&57.24$\pm$3.2&58.16$\pm$0.9&58.26$\pm$0.9&61.72$\pm$0.6&\textbf{61.76$\pm$0.5}\\
    &&40&57.08$\pm$3.2&56.05$\pm$2.0&61.01$\pm$1.3&62.71$\pm$0.7&53.96$\pm$1.1&63.74$\pm$2.7&57.82$\pm$0.8&54.28$\pm$1.6&64.03$\pm$0.7&\textbf{65.44$\pm$0.2}\\
    &&60&55.92$\pm$3.2&53.85$\pm$0.4&58.74$\pm$0.8&58.57$\pm$0.5&56.84$\pm$0.7&61.06$\pm$2.0&57.96$\pm$0.7&56.69$\pm$1.2&63.61$\pm$1.6&\textbf{64.65$\pm$0.5}\\
    \cline{2-13}
    &\multirow{3}{*}{AUC}
    &20&67.63$\pm$5.0&73.03$\pm$0.7&71.78$\pm$0.7&73.89$\pm$0.4&70.84$\pm$0.7&73.26$\pm$2.1&72.80$\pm$0.6&73.19$\pm$1.2&76.22$\pm$0.8&\textbf{76.40$\pm$0.1}\\
    &&40&66.42$\pm$4.7&74.05$\pm$0.9&75.51$\pm$0.8&76.08$\pm$0.4&69.48$\pm$0.2&77.74$\pm$1.2&72.97$\pm$1.1&70.77$\pm$1.6&78.44$\pm$0.5&\textbf{78.48$\pm$0.7}\\
    &&60&66.78$\pm$3.5&71.75$\pm$0.4&74.78$\pm$0.4&74.89$\pm$0.4&71.01$\pm$0.5&75.69$\pm$1.5&73.32$\pm$0.9&73.17$\pm$1.4&78.04$\pm$0.4&\textbf{78.13$\pm$0.4}\\
    \hline
    \multirow{9}{*}{AMiner}&
    \multirow{3}{*}{Ma-F1}
    &20&42.46$\pm$2.5&60.22$\pm$2.0&54.78$\pm$0.5&58.32$\pm$1.1&50.06$\pm$0.9&56.07$\pm$3.2&51.61$\pm$3.2&59.50$\pm$2.1&71.38$\pm$1.1&\textbf{76.34$\pm$0.2}\\
    &&40&45.77$\pm$1.5&65.66$\pm$1.5&64.77$\pm$0.5&64.50$\pm$0.7&58.97$\pm$0.9&63.85$\pm$1.5&54.72$\pm$2.6&61.92$\pm$2.1&73.75$\pm$0.5&\textbf{80.40$\pm$0.6}\\
    &&60&44.91$\pm$2.0&63.74$\pm$1.6&60.65$\pm$0.3&65.53$\pm$0.7&57.34$\pm$1.4&62.02$\pm$1.2&55.45$\pm$2.4&61.15$\pm$2.5&75.80$\pm$1.8&\textbf{80.22$\pm$0.1}\\
    \cline{2-13}
    &\multirow{3}{*}{Mi-F1}
    &20&49.68$\pm$3.1&65.78$\pm$2.9&60.82$\pm$0.4&63.64$\pm$1.1&61.49$\pm$2.5&68.86$\pm$4.6&62.39$\pm$3.9&63.93$\pm$3.3&78.81$\pm$1.3&\textbf{83.08$\pm$0.2}\\
    &&40&52.10$\pm$2.2&71.34$\pm$1.8&69.66$\pm$0.6&71.57$\pm$0.7&68.47$\pm$2.2&76.89$\pm$1.6&63.87$\pm$2.9&63.60$\pm$2.5&80.53$\pm$0.7&\textbf{86.72$\pm$0.9}\\
    &&60&51.36$\pm$2.2&67.70$\pm$1.9&63.92$\pm$0.5&69.76$\pm$0.8&65.61$\pm$2.2&74.73$\pm$1.4&63.10$\pm$3.0&62.51$\pm$2.6&82.46$\pm$1.4&\textbf{86.32$\pm$0.1}\\
    \cline{2-13}
    &\multirow{3}{*}{AUC}
    &20&70.86$\pm$2.5&85.39$\pm$1.0&81.22$\pm$0.3&83.35$\pm$0.5&77.96$\pm$1.4&78.92$\pm$2.3&75.89$\pm$2.2&85.34$\pm$0.9&90.82$\pm$0.6&\textbf{93.09$\pm$0.8}\\
    &&40&74.44$\pm$1.3&88.29$\pm$1.0&88.82$\pm$0.2&88.70$\pm$0.4&83.14$\pm$1.6&80.72$\pm$2.1&77.86$\pm$2.1&88.02$\pm$1.3&92.11$\pm$0.6&\textbf{93.86$\pm$0.3}\\
    &&60&74.16$\pm$1.3&86.92$\pm$0.8&85.57$\pm$0.2&87.74$\pm$0.5&84.77$\pm$0.9&80.39$\pm$1.5&77.21$\pm$1.4&86.20$\pm$1.7&92.40$\pm$0.7&\textbf{95.28$\pm$0.1}\\
    \hline
  \end{tabular}}
\end{table*}

\begin{figure*}[t]
\centering
\captionsetup{justification=centering}
\subfigure[Mp2vec]{
\label{Mp2vec_acm}
\includegraphics[width=0.15\textwidth]{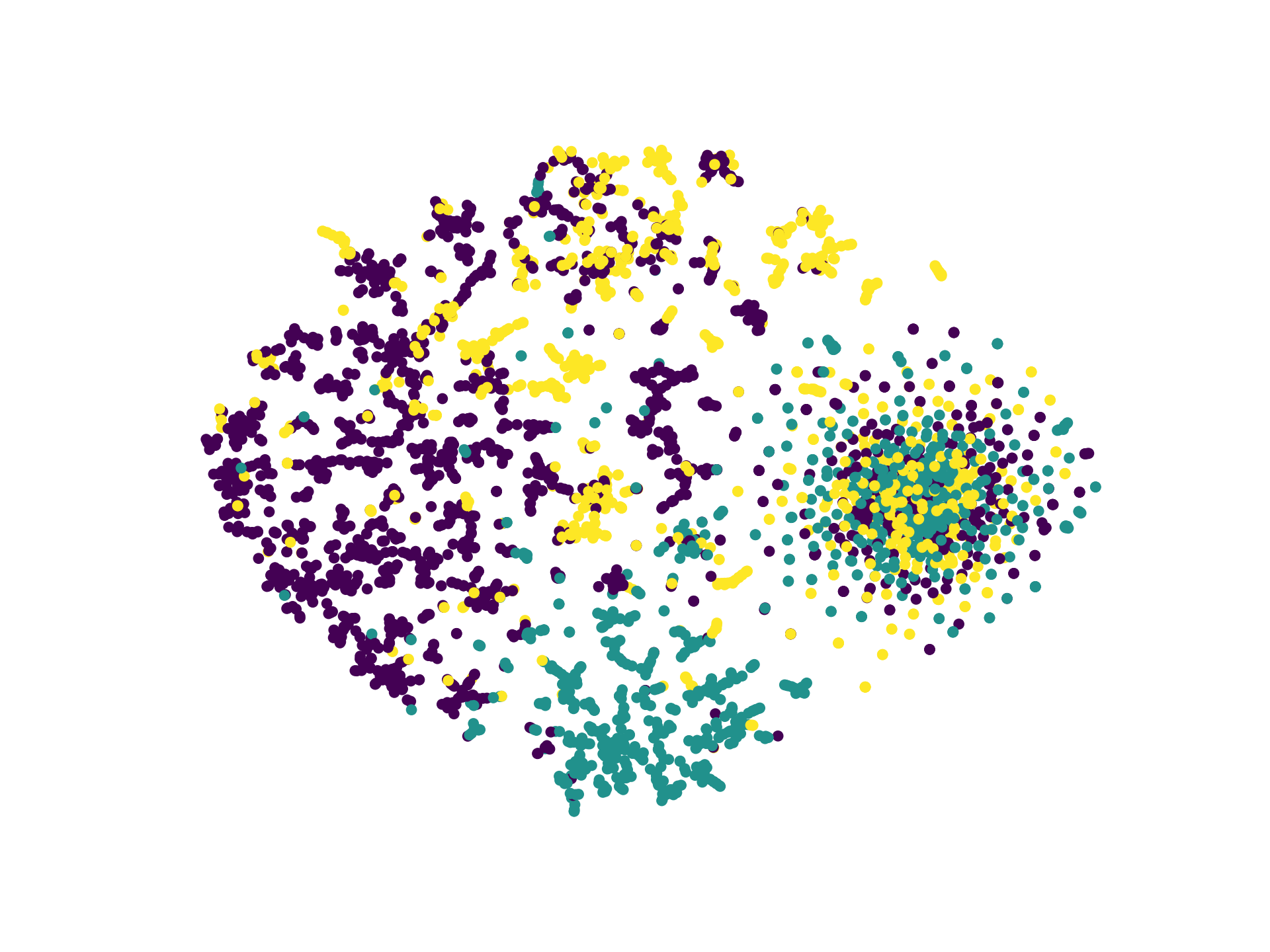}}
\subfigure[DGI]{
\label{DGI_acm}
\includegraphics[width=0.15\textwidth]{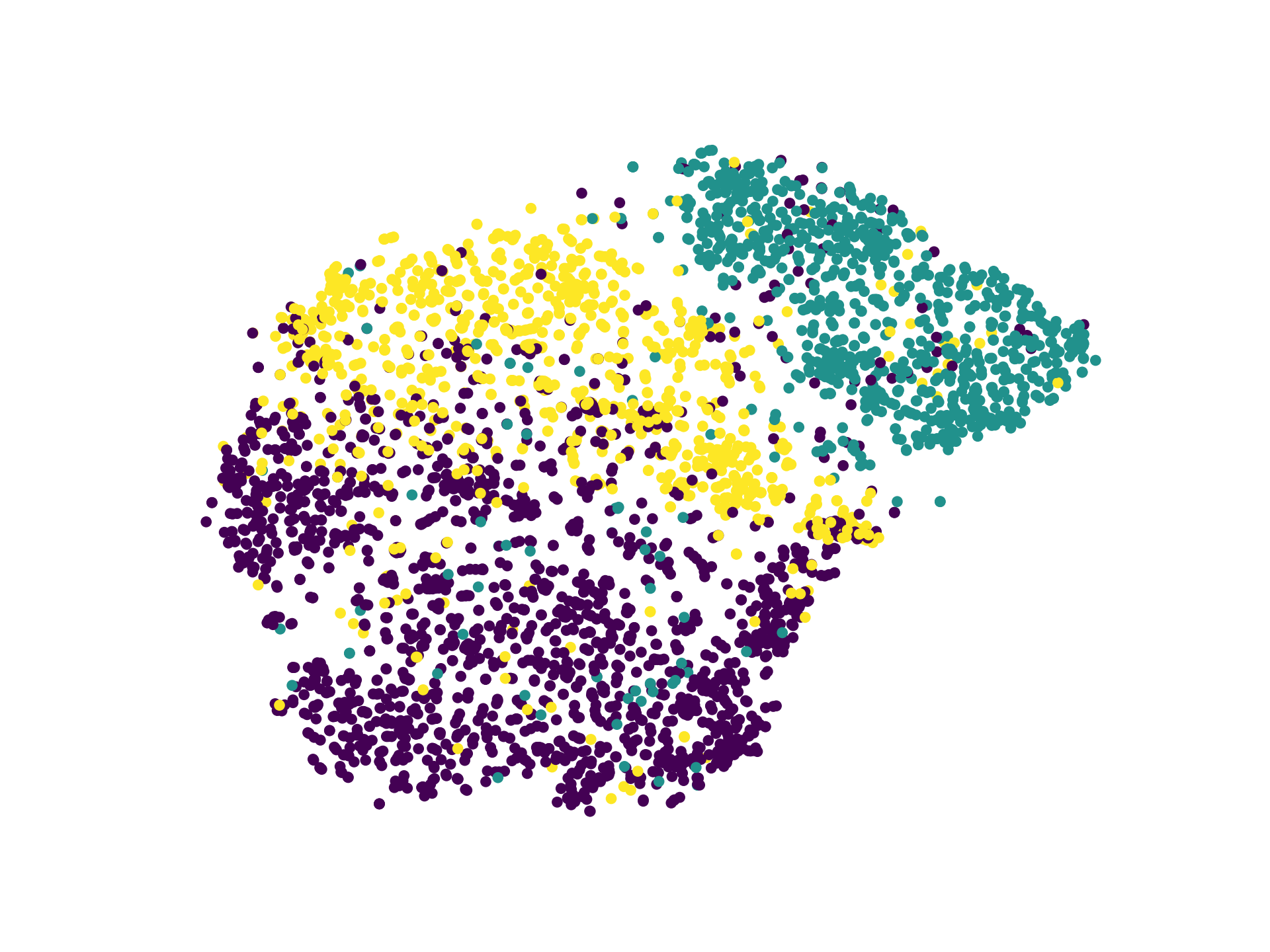}}
\subfigure[DMGI]{
\label{DMGI_v_acm}
\includegraphics[width=0.15\textwidth]{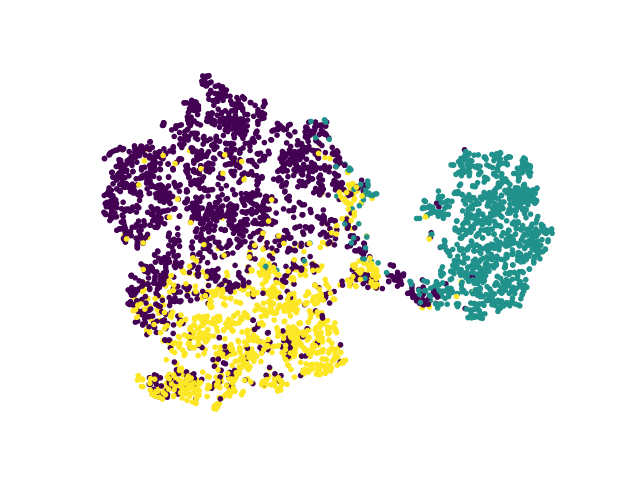}}
\subfigure[HeCo]{
\label{HeCo_v_acm}
\includegraphics[width=0.15\textwidth]{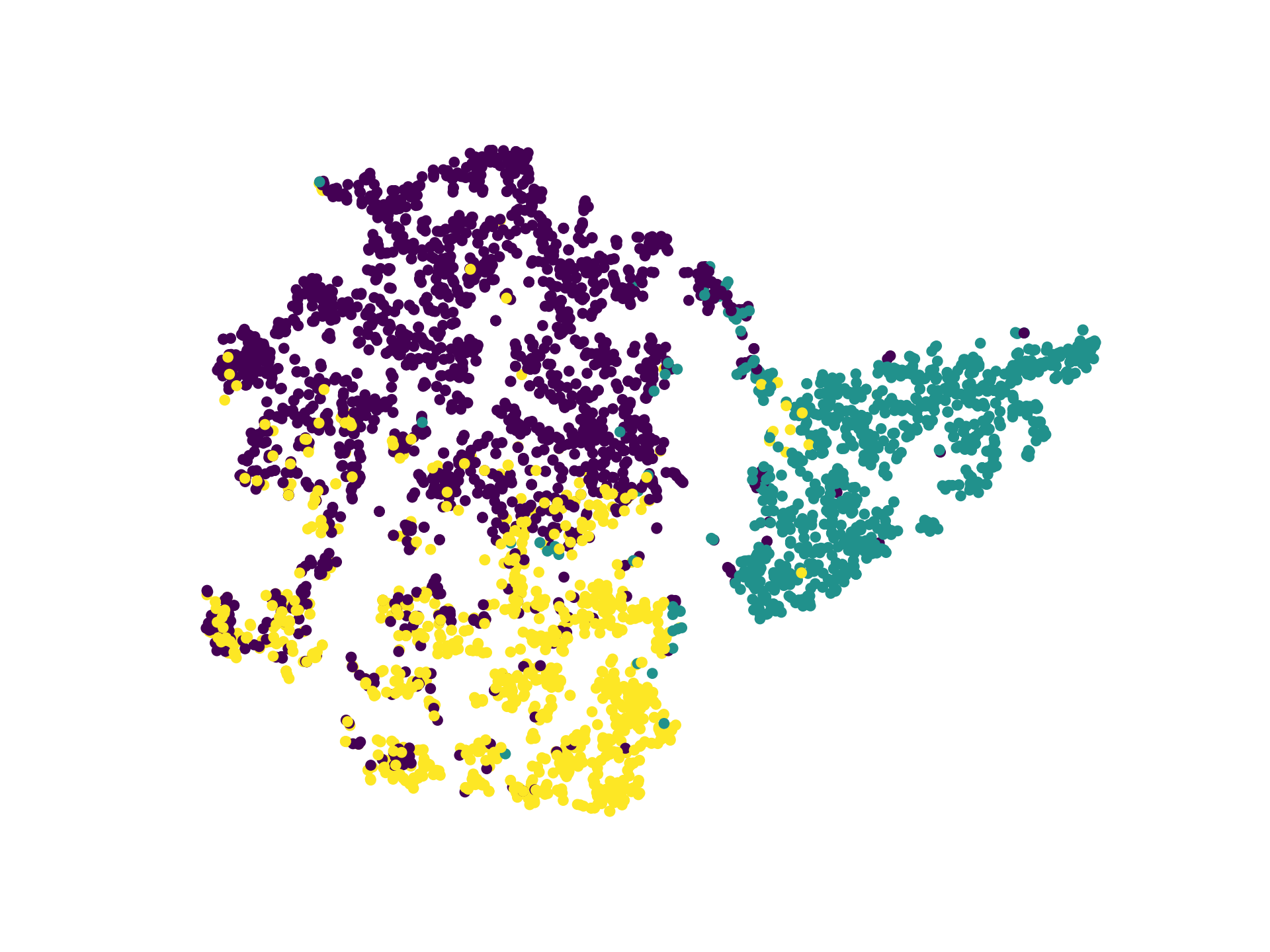}}
\subfigure[CSGRL]{
\label{CSGRL_v_acm}
\includegraphics[width=0.15\textwidth]{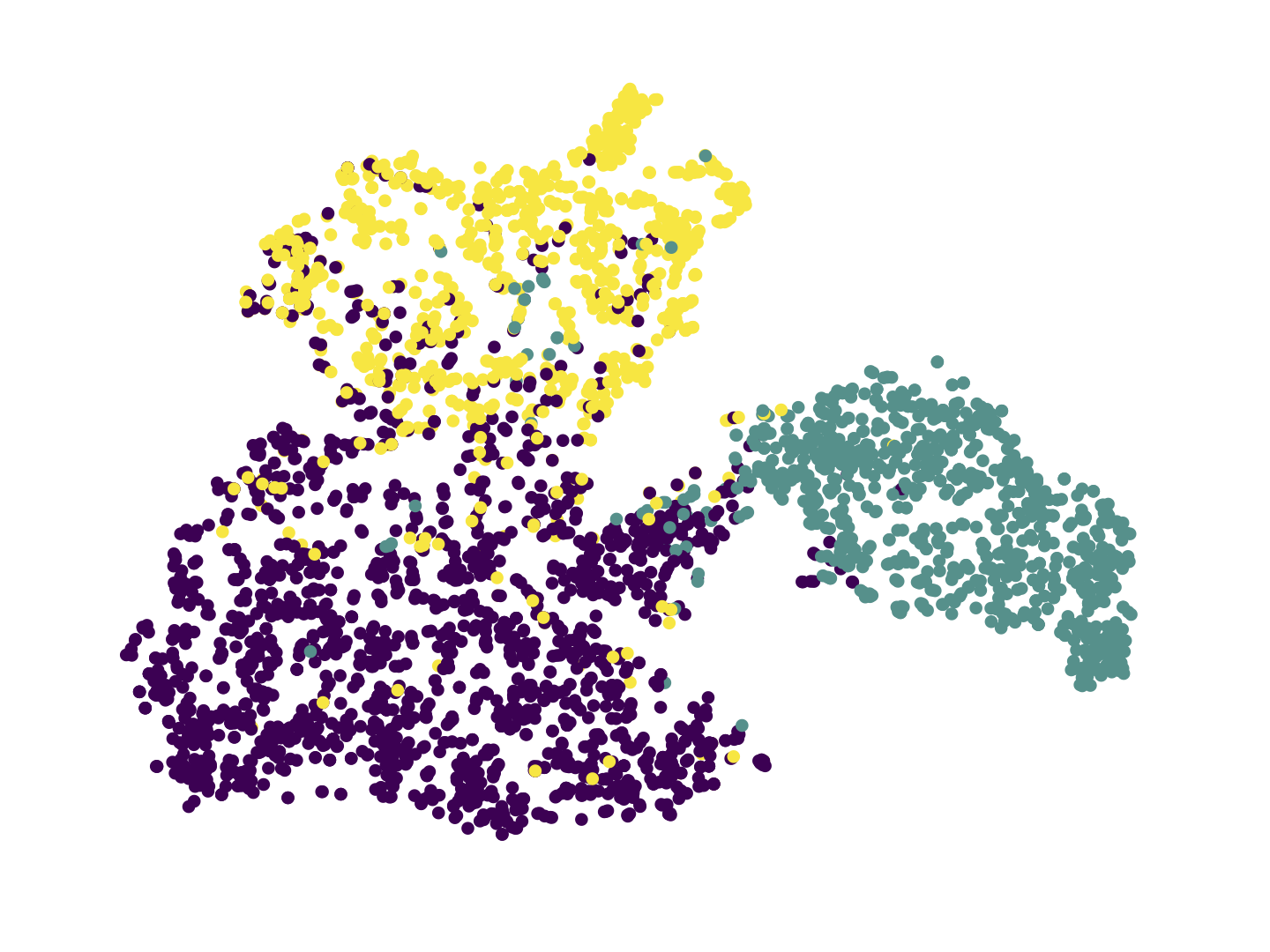}}
\caption{Visualization of the learned node embedding on ACM. The Silhouette scores for (a) (b) (c) (d) (e) are 0.0292, 0.1862, 0.3015 and 0.3642, 0.3782 respectively.}
\label{fig:visiulization}
\end{figure*}

\subsubsection{Datasets}
We use four real datasets. ACM\cite{nshe} is an academic network. The object of classification is papers, and there are three classes. DBLP\cite{magnn} is also an academic network. The object of classification is the paper, and there are four classes. Freebase\cite{freebase} is a film information network. The object of classification is movies and is divided into three classes. AMiner\cite{hegan} is an academic network, divided into four classes.

\subsubsection{Baselines}
The baseline consists of three unsupervised homogeneous methods (GraphSAGE\cite{GraphSAGE}, GAE\cite{GAE}, DGI\cite{dgi}) and five unsupervised heterogeneous methods ( Mp2vec\cite{mp2vec}, HERec\cite{herec}, HetGNN\cite{hetegnn}, DMGI\cite{dmgi}, HeCo\cite{heco}) and one semi-supervised heterogeneous method, HAN\cite{han}.

\subsubsection{Implementation Detail}
For random walk-based methods, the number of walks is set to $40$, the length is $100$, and the window size is set to $5$. For other parameters, follow the settings of the original document. Experimental results refer to those of HeCo. For CSGRL, the learning rate is set to $10^{-2}$ and weight decay is set to $10^{-5}$. To perform mini-batch training, HGT mini-batch method is used. The batch size is set to $256$. The optimization function uses AdamW. It is set to $0.99$ for $\tau$ in BYOL.

\begin{table}
  \caption{Quantitative results (\%$\pm\sigma$) on node clustering.}
  \label{tab:result2}
  \resizebox{0.47\textwidth}{!}{
  \begin{tabular}{c|cc|cc|cc|cc}
    \hline
    Datasets & \multicolumn{2}{c|}{ACM} & \multicolumn{2}{c|}{DBLP} & \multicolumn{2}{c|}{Freebase} & \multicolumn{2}{c}{AMiner} \\
    \hline
    Metrics & NMI & ARI & NMI & ARI & NMI & ARI & NMI & ARI\\
    \hline
    GraphSage&29.20&27.72&51.50&36.40&9.05&10.49&15.74&10.10\\
    GAE&27.42&24.49&72.59&77.31&19.03&14.10&28.58&20.90\\
    Mp2vec&48.43&34.65&73.55&77.70&16.47&17.32&30.80&25.26\\
    HERec&47.54&35.67&70.21&73.99&19.76&19.36&27.82&20.16\\
    HetGNN&41.53&34.81&69.79&75.34&12.25&15.01&21.46&26.60\\
    DGI&51.73&41.16&59.23&61.85&18.34&11.29&22.06&15.93\\
    DMGI&51.66&46.64&70.06&75.46&16.98&16.91&19.24&20.09\\
    HeCo&56.87&56.94&74.51&80.17&20.38&20.98&32.26&28.64\\
    \hline
    CSGRL&\textbf{66.13}&\textbf{70.84}&\textbf{74.88}&\textbf{80.24}&\textbf{22.76}&\textbf{22.81}&\textbf{44.45}&\textbf{39.25}\\
    \hline
  \end{tabular}}
\end{table}

\begin{table}
  \caption{Quantitative results (\%$\pm\sigma$) on node classification.}
  \label{tab:result3}
  \resizebox{0.47\textwidth}{!}{
  \begin{tabular}{c|ccc|ccc}
    \hline
    Datasets & \multicolumn{3}{c|}{ACM} & \multicolumn{3}{c}{AMiner} \\
    \hline
    Metrics & Ma-F1 & Mi-F1 & AUC & Ma-F1 & Mi-F1 & AUC\\
    \hline
    BGRL&90.43$\pm$0.2&90.26$\pm$0.2&97.05$\pm$0.1&64.28$\pm$0.9&70.36$\pm$1.1&87.90$\pm$0.3\\
    CSGRL-wopp&90.51$\pm$0.3&90.37$\pm$0.3&\textbf{97.15$\pm$0.1}&73.54$\pm$0.4&80.65$\pm$0.4&92.48±0.1\\
    CSGRL&\textbf{90.92$\pm$0.5}&\textbf{90.78$\pm$0.5}&97.08$\pm$0.5&\textbf{80.22$\pm$0.1}&\textbf{86.32$\pm$0.1}&\textbf{95.28$\pm$0.1}\\
    \hline
  \end{tabular}}
\end{table}

\subsection{Node Classification}
The learned embeddings are used to train the linear classifier. Assuming there are 20, 40, or 60 labeled nodes for each class as training set. For each dataset, 1000 is set as the validation set and 1000 as the test set. We use common evaluation metrics such as Macro-F1, Micro-F1, AUC. Referring to Table~\ref{tab:result1}, In all cases, our model outperformed other competing models, indicating that the BYOL approach is effective for unsupervised learning.

\subsection{Node Clustering}
In this task, we compute and report the NMI and ARI using the K-means algorithm for learned embeddings. Referring to Table~\ref{tab:result2}, It also outperforms other competing models in node clustering.

\subsection{Node Visualization}
To provide a more intuitive assessment, we provide embedding visualizations for ACM. Comparison with Mp2vec, DGI, DMGI, HeCo using T-SNE. In Figure~\ref{fig:visiulization}, embeddings are clearly separated by class, and the silhouette score is higher than that of other comparative models, indicating that embeddings were successful.

\subsection{Analysis with CSGRL variant}
Referring to Table~\ref{tab:result3}, it can be seen that the performance of BGRL and CSGRL is similar in the case of ACM, which is a dataset with attributes, but the performance of BGRL is not good in the case of AMiner, which is a non-attribute dataset. For CSGRL-wopp, a model trained without positive pairs, we can observe that the performance is comparable to or worse than CSGRL. Here, we prove that using a positive pair is effective for improving performance.

\section{Conclusion}
In this paper, we propose a self-supervised heterogeneous graph neural network called CSGRL using a BYOL methodology and enhanced augmentation methods. CSGRL learns how to efficiently represent complex information using two views: network schema and meta-path. We also use the BYOL methodology to avoid the problem of finding positive and negative pairs, one of the major difficulties that exist in self-supervised heterogeneous graph learning. In extensive experiments, it has been proven that our model outperforms other models.

\bibliographystyle{IEEEtran}
\bibliography{IEEEabrv,./ref}
\end{document}